\documentclass[10pt,twocolumn,letterpaper]{article}

\usepackage[pagenumbers]{cvpr} % To force page numbers, e.g. for an arXiv version

\usepackage{graphicx}
\usepackage{amsmath}
\usepackage{amssymb}
\usepackage{booktabs}
\usepackage{multirow}
\usepackage{bm}
\usepackage[misc]{ifsym}
\usepackage{pifont}
\usepackage{lipsum}
\usepackage{subfloat}
\usepackage[pagebackref,breaklinks,colorlinks]{hyperref}
\usepackage{bbding}
\newcommand{\first}[1]{{\color{red} \textbf{#1}}}
\newcommand{\second}[1]{{\color{blue} \underline{#1}}}

\usepackage[capitalize]{cleveref}

\crefname{section}{Sec.}{Secs.}
\Crefname{section}{Section}{Sections}
\Crefname{table}{Table}{Tables}
\crefname{table}{Tab.}{Tabs.}

\newcommand\blfootnote[1]{%
\begingroup
\renewcommand\thefootnote{}\footnote{#1}%
\addtocounter{footnote}{-1}%
\endgroup
}
\begin{document}

%%%%%%%%% TITLE - PLEASE UPDATE
\title{Extracting Motion and Appearance via Inter-Frame Attention \\for Efficient Video Frame Interpolation}

\author{\hspace{-5.5mm} Guozhen Zhang\textsuperscript{1} \quad Yuhan Zhu\textsuperscript{1} \quad Haonan Wang\textsuperscript{1} \quad  Youxin Chen\textsuperscript{3} \quad  Gangshan Wu\textsuperscript{1} \quad  Limin Wang\textsuperscript{1, 2,~\Letter} \\
\textsuperscript{1}State Key Laboratory for Novel Software Technology, Nanjing University, China \\
\textsuperscript{2}Shanghai AI Lab, China \quad
\textsuperscript{3}Samsung Electronics (China) R\&D Centre, China
}
\maketitle

%%%%%%%%% ABSTRACT
\begin{abstract}
   Effectively extracting inter-frame motion and appearance information is important for video frame interpolation (VFI). Previous works either extract both types of information in a mixed way or elaborate separate modules for each type of information, which lead to representation ambiguity and low efficiency. In this paper, we propose a novel module to explicitly extract motion and appearance information via a unifying operation. Specifically, we rethink the information process in inter-frame attention and reuse its attention map for both appearance feature enhancement and motion information extraction. Furthermore, for efficient VFI, our proposed module could be seamlessly integrated into a hybrid CNN and Transformer architecture. This hybrid pipeline can alleviate the computational complexity of inter-frame attention as well as preserve detailed low-level structure information. Experimental results demonstrate that, for both fixed- and arbitrary-timestep interpolation, our method achieves state-of-the-art performance on various datasets. Meanwhile, our approach enjoys a lighter computation overhead over models with close performance. The source code and models are available at \url{https://github.com/MCG-NJU/EMA-VFI}.
\end{abstract}
\blfootnote{ \Letter: Corresponding author (lmwang@nju.edu.cn).}
% be available at \href{.}{.}

%%%%%%%%% BODY TEXT
\section{Introduction}
\label{sec:intro}
% 	As a fundamental low-level vision task, the goal of video frame interpolation~(VFI) is to generate intermediate frames given a pair of consecutive frames. It has a wide range of real-life applications, such as video compression~\cite{wu2018video}, novel-view rending~\cite{szeliski1999prediction,flynn2016deepstereo}, and creating slow-motion video~\cite{jiang2018super}. To synthesize intermediate frames between consecutive frames, the video frame interpolation method needs to capture the motion between the input frames and fuse the corresponding appearance of the input frames into the intermediate frames, that is to say, it is necessary to extract the motion and appearance information between input frames.
    As a fundamental low-level vision task, the goal of video frame interpolation~(VFI) is to generate intermediate frames given a pair of consecutive frames~\cite{huang2020rife,lu2022video}. It has a wide range of real-life applications, such as video compression~\cite{wu2018video}, novel-view rending~\cite{szeliski1999prediction,flynn2016deepstereo}, and creating slow-motion video~\cite{jiang2018super}. In general, VFI can be seen as the process of capturing the motion between consecutive frames and then blending the corresponding appearance to synthesize the intermediate frames. From this perspective, the motion and appearance information between input frames are important for achieving excellent performance in the VFI task.
 	\begin{figure}[t]
		\begin{center}
			%\fbox{\rule{0pt}{2in} \rule{0.9\linewidth}{0pt}}
			\includegraphics[width=1.0\linewidth]{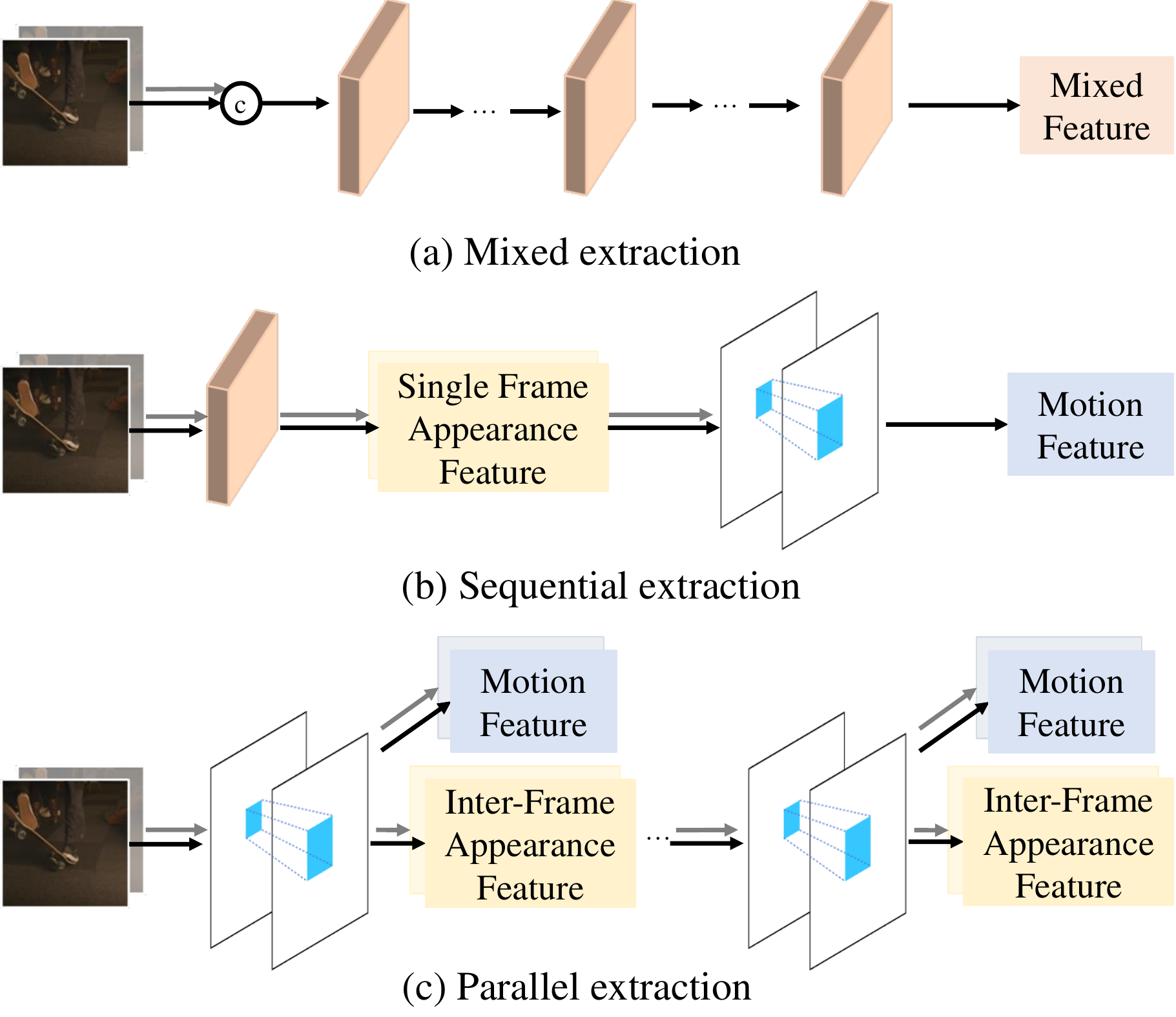}
		\end{center}
		\vspace{-0.13in}
		\caption{Illustration of various approaches in video frame interpolation for acquiring motion  and appearance information. }
		\label{fig:intro}
	\end{figure}

        Concerning the extraction paradigm of motion and appearance information, the current VFI approaches can be divided into two categories. The first is to handle both appearance and motion information in a mixed way~\cite{shi2022video, niklaus2017video, niklaus2017videoas, ding2021cdfi, gui2020featureflow,kalluri2020flavr,long2016learning,lu2022video,huang2020rife,bao2019depth,kong2022ifrnet}, as shown in \cref{fig:intro}(a). The two neighboring frames are directly concatenated and fed into a backbone composed of successively similar modules to generate features with mixed motion and appearance information. Though simple, this approach requires an elaborate design and high capacity in the extractor module, as it needs to deal with both motion and appearance information jointly. The absence of explicit motion information also results in limitations for arbitrary-timestep interpolation. The second category, as shown in \cref{fig:intro}(b), is to design separate modules for motion and appearance information extraction~\cite{reda2022film,niklaus2018context,danier2022st,xue2019video,sim2021xvfi,jia2022neighbor, park2020bmbc,park2021asymmetric}. This approach requires additional modules, such as cost volume~\cite{jia2022neighbor,park2020bmbc,park2021asymmetric}, to extract motion information, which often imposes a high computational overhead. Also, only extracting appearance features from a single frame fails to capture the correspondence of appearance information of the same regions between frames, which is an effective cue for the VFI task~\cite{jia2022neighbor}.
        % The above observations lead us to wonder: is there a module that can simultaneously extract motion and appearance features in a distinguishable and unified way?

        To address the issues of the above two extraction paradigms, in this paper, we propose to explicitly extract both motion and appearance information via a unifying operation of inter-frame attention.
        With single inter-frame attention, as shown in \cref{fig:intro}(c), we are able to enhance the appearance features between consecutive frames and acquire motion features at the same time by reusing the attention maps. This basic processing unit could be stacked to obtain the hierarchical motion and appearance information. 
       Specifically, for any patch in the current frame, we take it as the query and its temporal neighbors as keys and values to derive an attention map representing their temporal correlation. After that, the attention map is leveraged to aggregate the appearance features of neighbors to contextualize the current region representation. In addition, the attention map is also used to weight the displacement of neighbors to get an approximate motion vector of the patch from the current frame to the neighbor frame. Finally, the obtained features are utilized with light networks for motion estimation and appearance refinement to synthesize intermediate frames.
       Compared with previous works, our design enjoys three merits. (1) The appearance features of each frame can be enhanced with each other yet not be mixed with motion features to preserve the detailed static structure information. (2) The obtained motion features can be scaled by time and then used as cues to guide the generation of frames at any moment between input frames.  (3) We only need to control the complexity and the number of modules to balance the overall performance and the inference speed.

        Directly using inter-frame attention on original resolution results in huge memory usage and computational overhead. Inspired by some recent works~\cite{wang2022convolutional,d2021convit,dai2021coatnet,li2021localvit,wu2021cvt,xiao2021early,yuan2021incorporating}, which combines Convolutional Neural Network~(CNN) with Transformer~\cite{vaswani2017attention} to improve the model learning ability and robustness, we adopt a simple but effective architecture: utilize CNN to extract high-resolution low-level features and then use Transformer blocks equipped with inter-frame attention to extracting low-resolution motion features and inter-frame appearance features. Our proposed module could be seamlessly integrated into this hybrid pipeline to extract motion and appearance features efficiently without losing fine-grained information. In summary, our contributions include: 
        \begin{itemize}
            \item We propose to utilize inter-frame attention to extract both motion and appearance information simultaneously for video frame interpolation.
            \item An hybrid CNN and Transformer design is adopted to overcome the overhead bottleneck of the inter-frame attention at high-resolution input while preserving fine-grained information.
            \item Our model achieves state-of-the-art performance on various datasets while being efficient compared to models with nearby performance.
        \end{itemize}
        % (1) We propose to utilize inter-frame attention to extract both motion and appearance information simultaneously for video frame interpolation. (2) An integrated CNN and Transformer design is adopted to overcome the overhead bottleneck of the inter-frame attention at high-resolution input while preserving fine-grained information. (3) Our model achieves state-of-the-art performance on various datasets while being simple and fast compared to models with nearby performance.

%------------------------------------------------------------------------
\section{Related Work}
	
	\subsection{Video Frame Interpolation}
        The current VFI methods can be roughly divided into two categories: mixed methods and motion-aware methods.
        %------------------------------------------------------------------------
        \textbf{Mixed methods} tends to generate intermediate frames by directly concatenating input frames and feeding into a feature backbone to handle both motion and appearance without explicitly motion representation. In terms of how the intermediate frames are obtained, this method can be subdivided into two categories: directly-generated methods and kernel-based methods. Directly-generated methods~\cite{gui2020featureflow,long2016learning,choi2020channel,kalluri2020flavr} generate intermediate frames directly end-to-end from the input frames. Kernel-based methods~\cite{shi2022video,lee2020adacof,cheng2020video,cheng2021multiple,niklaus2017video,niklaus2017videoas,niklaus2021revisiting,ding2021cdfi} generate interpolated frames by learning kernels and performing local convolution on the input frames. Although these methods are relatively simple, their lack of modeling of motion makes it difficult to match the corresponding regions between intermediate frames and input frames, leading to image blur and artifacts~\cite{lee2022enhanced}.
        %------------------------------------------------------------------------
        \textbf{Motion-aware methods} explicitly model the motion (usually represented by optical flow) between two frames to assist in aligning the appearance information of the input frames to intermediate frames. Some early work~\cite{liu2017video,jiang2018super,liu2019deep} did not exploit the appearance information of the input frames and only predicted inter-frame motion for pixel-level alignment. Niklaus~\etal~\cite{niklaus2018context} first proposes to refine the aligned intermediate frames with a synthesis network utilizing the contextual features. Most of the following works ~\cite{niklaus2018context,danier2022st,xue2019video,kong2022ifrnet,bao2019depth,niklaus2020softmax,park2020bmbc,huang2020rife,sim2021xvfi,park2021asymmetric,lu2022video,reda2022film} design separate modules for explicitly motion modeling and appearance synthesis to boost the performance. Though the current state-of-the-art method~\cite{lu2022video} has achieved surprising performance, the increasing system complexity makes it unrealistic to apply in practice. Our proposed method also explicitly models the motion but could extract motion and appearance information in a unified and efficient way.
        % \begin{figure*}[t]
        % \begin{center}
        % \includegraphics[width=0.79\linewidth]{imgs/Method1.pdf}
        % \end{center}
        % \vspace{-0.15in}
        % \caption{(a) An example of how inter-frame attention acquires motion and inter-frame appearance features. For any region in $I_0$, we use it as a query and the spatial neighbors in $I_1$ as keys/values to generate an attention map. Then we exploit the attention map to aggregate the appearance information in $I_1$ to get an inter-frame appearance representation of the query region, and meanwhile, estimate an approximate displacement of the query region between frames. (b) An illustration of Transformer blocks employing inter-frame attention. We basically follow the conventional design as \cite{vaswani2017attention} while maintaining the spatial-temporal structure of different frames.}
        % 		\label{fig:att}
        %             \vspace{-0.2in}
        % 	\end{figure*}
        \begin{figure*}[t]
        \centering
        \subfloat[]{\includegraphics[width=0.46\linewidth]{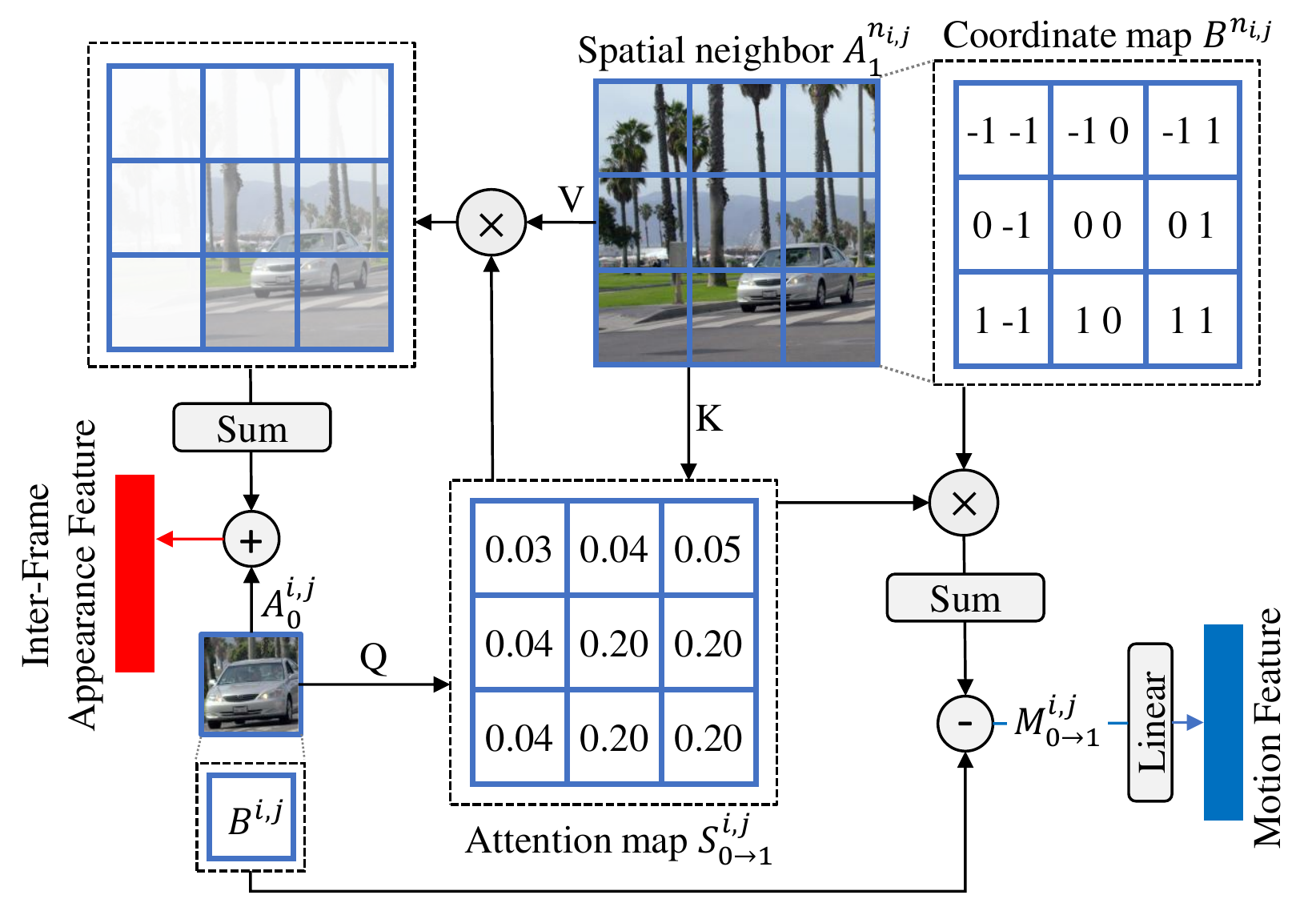}\label{fig:ifa}}
        \quad \quad \quad \quad 
        \subfloat[]{\includegraphics[width=0.23\linewidth]{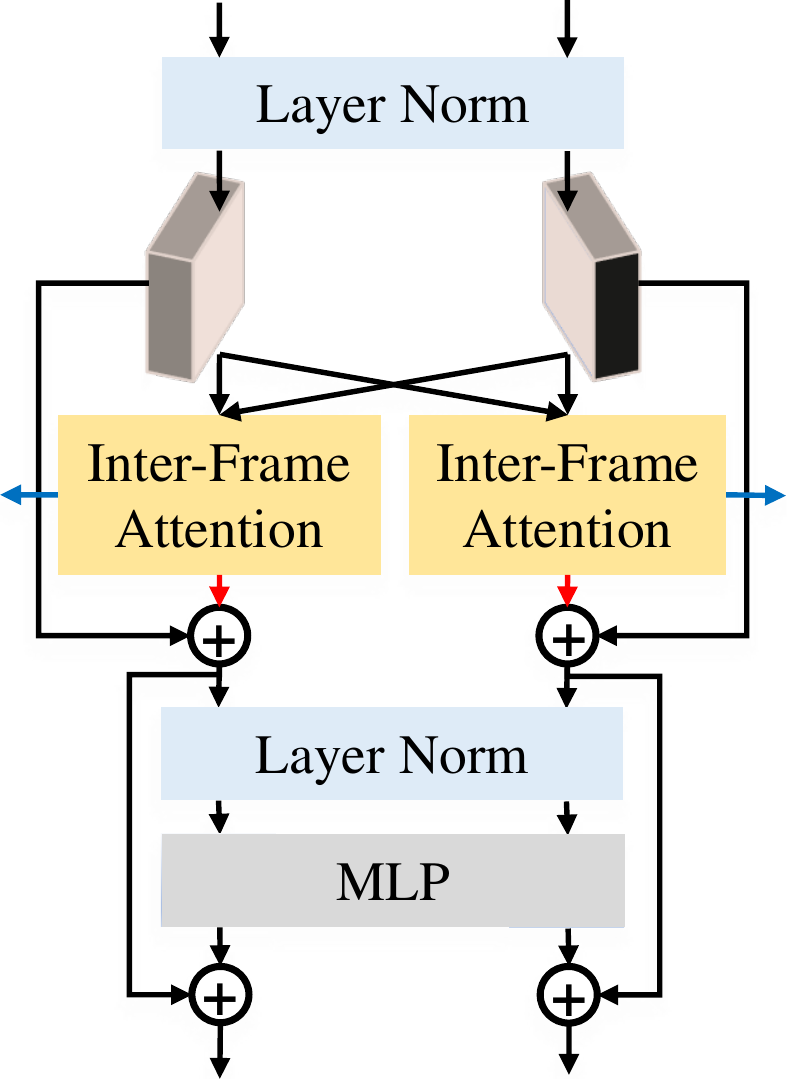}\label{fig:tf}}
        % \vspace{-0.15in}
        \caption{(a) An example of how inter-frame attention acquires motion and inter-frame appearance features. For any region in $I_0$, we use it as a query and the spatial neighbors in $I_1$ as keys/values to generate an attention map. Then we exploit the attention map to aggregate the appearance information in $I_1$ to get an inter-frame appearance representation of the query region, and meanwhile, estimate an approximate displacement of the query region between frames. (b) An illustration of Transformer blocks employing inter-frame attention. We basically follow the conventional design as \cite{vaswani2017attention} while maintaining the spatial-temporal structure of different frames.}
        \label{fig:att}
        \vspace{-0.2in}
        \end{figure*}
        %------------------------------------------------------------------------
        \subsection{Extracting Motion and Appearance}
        Although it has been rarely explored in the VFI task, there are a considerable number of articles in the video understanding discussing how to extract motion information and appearance information simultaneously~\cite{diba2018spatio,wang2018appearance,zhao2018recognize,kwon2020motionsqueeze,wang2020video}. Wang~\etal~\cite{wang2018appearance} exploits learnable multiplicative interactions to acquire relation between frames and fuse it with appearance to generate spatiotemporal features. Zhao~\etal~\cite{zhao2018recognize} derive disentangled components of dynamics purely from raw video frames, which comprise the static appearance, apparent motion, and appearance changes. Some following works~\cite{wang2020video,kwon2020motionsqueeze} have also improved this approach with more flexible and dynamic operations. The apparent motion in Zhao~\etal~\cite{zhao2018recognize} is conceptually the closest to the motion feature in our paper, which uses expected displacement at each point based on a distribution over correspondences to represent motion. Compared to these methods, we are the first to exploit inter-frame attention to extract motion and appearance information directly.
        %------------------------------------------------------------------------
        \subsection{Transformer}
        Transformer~\cite{vaswani2017attention} has recently been widely used in different tasks of computer vision, and recent works~\cite{shi2022video,lu2022video} also introduce this architecture into video frame interpolation to leverage the flexibility and ability to capture long-range correspondence. However, when interpolating frames for high-resolution videos, these methods require much more computation and memory overhead compared to models using CNN. Recently, some studies have shown that combining CNN with Transformers improves the performance of the model~\cite{wang2022convolutional,d2021convit,dai2021coatnet,li2021localvit,wu2021cvt,xiao2021early,yuan2021incorporating}. Inspired by these methods, our proposed model adopts a similar idea by first extracting high-resolution features using CNN and then using Transformers to capture the motion features and enhanced appearance features.
	
	%------------------------------------------------------------------------
	\section{Our Method}
 
	Our goal is to generate the frame $\hat{\bm{I}_t} \in \mathbb{R}^{H \times W \times 3 }$  at any arbitrary timestep $t\in (0,1)$ given frames $\bm{I}_0,\bm{I}_1 \in \mathbb{R}^{ H \times W \times 3}$ at timestep $t=0$ and $t=1$, as:
    \begin{equation}
        \hat{\bm{I}}_t = \mathcal{O}(\bm{I}_0,\bm{I}_1,t),
    \end{equation}
    where $\mathcal{O}$ is our model. In the following, we first present the process of how to exploit inter-frame attention to extract motion and inter-frame appearance features simultaneously for video frame interpolation and the structure of Transformer blocks equipped with inter-frame attention in \cref{sec:MAFormer}. Next, we give a detailed explanation of the overall pipeline which utilizes a fused CNN design to overcome the heavy overhead brought by Transformer blocks while maintaining the fine-grained features in \cref{sec:overall}.
    
        \begin{figure*}[t]
		\centering
			\includegraphics[width=0.75\linewidth]{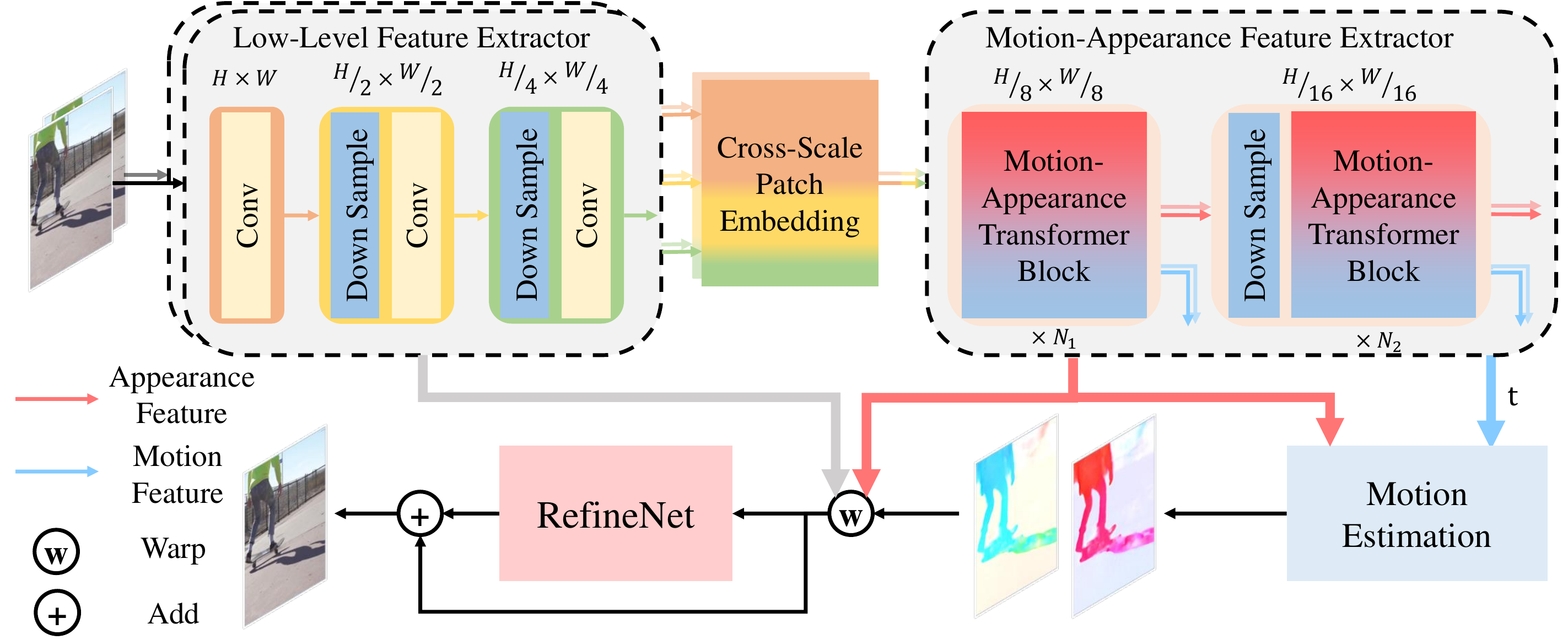}
		%\vspace{-0.18in}
		\caption{Overview of our proposed architecture. First, a low-level feature extractor composed of hierarchical convolutional layers is used to generate multi-scale fine-grained features and also reduce the input size of the Transformer for efficiency. These fine-grained features are then fused by a cross-scale path embedding for enhancing detailed information and fed into the proposed motion-appearance feature extractor to acquire motion and appearance features. Finally, the motion feature and the appearance feature are used for motion estimation and appearance refinement.}
		\label{fig:overall}
		\vspace{-0.25in}
	\end{figure*}
 
	\subsection{Extract Motion and Appearance Information}
        \label{sec:MAFormer}
        Capturing motion between input frames and fusing the inter-frame appearance features are critical to the VFI task. Previous methods either extract both information by directly concatenating frames and feeding into a feature backbone or elaborate complex modules respectively, \eg ContextNet~\cite{niklaus2018context,huang2020rife} for appearance and cost volume~\cite{park2020bmbc,park2021asymmetric} for motion. In contrast, we propose to utilize inter-frame attention to extract distinguishable motion and appearance information uniformly. Our motivation for using inter-frame lies in its ability to naturally model inter-frame motion and transfer appearance information at the same time.
        
        \noindent
        \textbf{Inter-frame Attention~(IFA).} An example of how inter-frame attention acquires motion and inter-frame appearance is shown in \cref{fig:ifa}. For the sake of brevity, here we only take the example of obtaining the motion and enhancing appearance information of $\bm{I}_0$. Now suppose we have the appearance feature of two frames, denoted as $\bm{A}_0$ and $\bm{A}_1 \in \mathbb{R}^{\hat H\times \hat W \times C}$. For any region, which is denoted as $\bm{A}_0^{i,j} \in \mathbb{R}^C$ in $\bm{I}_0$, we use it and its spatial neighbors $\bm{A}_1^{n_{i,j}} \in \mathbb{R}^{N\times N\times C}$ in $\bm{I}_1$, where $N$ represents the neighborhood window size, to generate the query and keys/values respectively: 
        \begin{align}
            &\bm{Q_0}^{i,j}=\bm{A}_0^{i,j}\bm{W}_Q\;,\\
            &\bm{K_1}^{n_{i,j}}=\bm{A}_1^{n_{i,j}}\bm{W}_K\;,\\ &\bm{V_1}^{n_{i,j}}=\bm{A}_1^{n_{i,j}}\bm{W}_{V}\;, 
        \end{align}
        where $\bm{W}_Q,\bm{W}_K,\bm{W}_V\in\mathbb{R}^{C\times\hat C}$ are linear projection matrices. Then we make a dot product between $\bm{Q_0}^{i,j}$ and each position of $\bm{K_1}^{n_{i,j}}$ and then apply SoftMax following \cite{vaswani2017attention} to generate the attention map $\bm{S}_{0\rightarrow 1}^{i,j} \in \mathbb{R}^{N\times N}$, where the value at each location represents the degree of similarity between $\bm{A}_0^{i,j}$ and its neighbors, as:
        \begin{align}
            \bm{S}_{0\rightarrow 1}^{i,j} = \text{SoftMax}\left(\bm{Q_0}^{i,j} \left(\bm{K_1}^{n_{i,j}}\right)^\mathrm{T}/ \sqrt{\hat{C}} \right)\;.
        \end{align}

        The obtained $\bm{S}_{0\rightarrow 1}^{i,j}$ can be utilized to transfer the appearance information and extract motion information simultaneously. As for appearance, we first aggregate the similar appearance information from $\bm{I}_1$ and then fuse it with $\bm{A}_0^{i,j}$ to enhance the appearance information in $\bm{I}_0$, as:
        \begin{align}
            \hat{\bm{A}}_0^{i,j} = \bm{A}_0^{i,j} + \bm{S}_{0\rightarrow 1}^{i,j}\bm{V_1}^{n_{i,j}}\;.
        \end{align}
        
        The enhanced appearance feature $\hat{\bm{A}}_0^{i,j}$ contains the blending of the appearance of the similar region in two different frames, which can provide more information on how the appearance is transformed between frames for generating intermediate frames.
        \begin{figure}[t]
		\begin{center}
			%\fbox{\rule{0pt}{2in} \rule{0.9\linewidth}{0pt}}
			\includegraphics[width=1\linewidth]{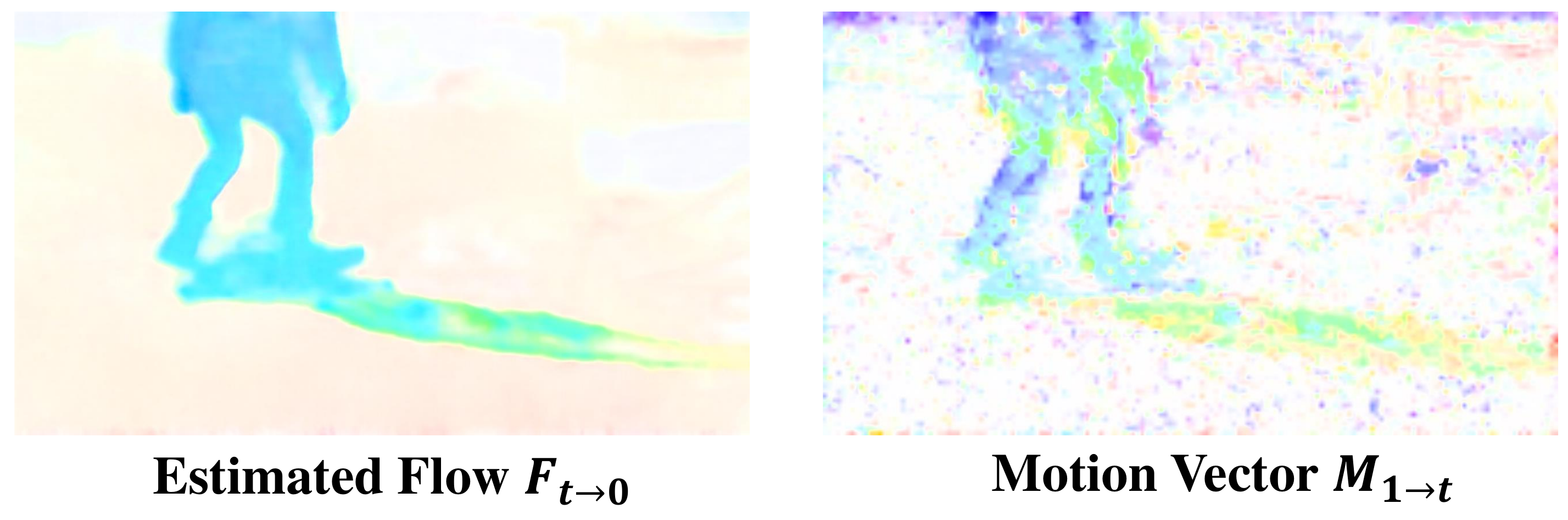}
		\end{center}
            \vspace{-0.2in}
		\caption{Visualization of estimated flow and motion vector.}
		\label{fig:mv}
		\vspace{-0.25in}
    \end{figure}
    
        As for motion, we first create a coordinate map $\bm{B} \in \mathbb{R}^{\hat{H}\times \hat{W} \times 2}$ in which the value at each location indicates the relative position in the entire image~((-1,-1) in the top-left and (1,1) in the bottom-right), as shown in \cref{fig:att}(a). Then we weight the neighbors' coordinates to estimate the approximately corresponding position of $\bm{A}_0^{i,j}$ in $\bm{I}_1$ and the motion vector $\bm{M}_{0\rightarrow 1}^{i,j} \in \mathbb{R}^{2}$ of $\bm{A}_0^{i,j}$ can be then generated by subtracting between the original position of $\bm{A}_0^{i,j}$ and the estimated position in $\bm{I}_1$, as:
        \begin{align}
            \bm{M}_{0\rightarrow 1}^{i,j} = \bm{S}_{0\rightarrow 1}^{i,j}\bm{B}^{n_{i,j}}- \bm{B}^{i,j}\;.
        \end{align}
        
        $\bm{M}_{0\rightarrow 1}^{i,j}$ contains motion information that can provide an explicit prior for motion estimation. The motion feature is then generated by passing $\bm{M}_{0\rightarrow 1}^{i,j}$ through a linear layer. It is worth noting that under the assumption of local linear motion, we can approximate the motion vector from $\bm{I}_0$ to $\bm{I}_t$ by multiplying $\bm{M}_{0\rightarrow 1}^{i,j}$ with t, as:
        \begin{align}
            \bm{M}_{0\rightarrow t}^{i,j} = t \times \bm{M}_{0\rightarrow 1}^{i,j}\;.
        \end{align}
        
        In this way, $\bm{M}_{0\rightarrow t}^{i,j}$ can be used as cues to guide the following motion estimation for arbitrary timestep frame prediction with only calculating $\bm{M}_{0\rightarrow 1}^{i,j}$ once. Note that the appearance feature $\hat{\bm{A}}_0^{i,j}$ is also timestep-invariant and hence the inter-frame attention only needs to be calculated once for multiple arbitrary timestep frame predictions. 
        
        \noindent
        \textbf{Discussion.} To demonstrate that the similarity of the same regions between frames can be captured by inter-frame attention, we compare the optical flow estimated by our trained model with the obtained motion vector. As shown in \cref{fig:mv}, motion vectors indeed maintain a high degree of consistency with predicted optical flow despite the presence of minor noise, which implies that IFA does have the ability to discriminate different regions and $\bm{M}_t$ can provide a strong prior for motion estimation. More quantitative support are provided in \cref{exp:ab}.
        
        \noindent
        \textbf{Structure of Transformer blocks.} We put the inter-frame attention within the Transformer block because it 
          has been proven to be effective in many vision tasks. As in \cref{fig:tf}, we basically followed the original Transformer design~\cite{vaswani2017attention} but modified it for the VFI task in two points: (1) We maintain the spatial-temporal structure of the different frames to perform IFA for extracting distinguishable features. (2) To accommodate different sizes of input frames and enhance the interaction between different regions in the same frame, we perform a similar strategy to \cite{chu2021conditional, wang2022pvt}, in which we remove the original position encoding and replace it with a depth-wise convolution in the multi-layer perceptron.

        \subsection{Overall Pipeline}
        \label{sec:overall}
        Our overall pipeline is illustrated in ~\cref{fig:overall}. Since the resolution of input frames can be very high, directly performing inter-frame attention on the original size brings huge memory usage and computation overhead. Inspired by some recent works~\cite{wang2022convolutional,xiao2021early,wu2021cvt}, we first utilize hierarchical convolutional layers as the low-level feature extractor to generate multi-scale single-frame  appearance features, as:
        \begin{align}
            &\bm{L}_i^0, \bm{L}_i^1, \bm{L}_i^2 = \mathcal{F}(\bm{I}_i)\;,
        \end{align}
        where $\mathcal{F}$ represents the low-level feature extractor and $\bm{L}_i^{k}$ represents the appearance feature of $i$-th frame with the shape $\frac{H}{2^k}\times \frac{W}{2^k}\times 2^k C$. The number of channels $C$ would be doubled each time the feature size reduces. Though this hybrid CNN and Transformers design could relieve the overhead, it also caused a lack of fine-grained information in the input of Transformers. To alleviate this problem, we reuse the low-level features derived by CNNs to complement the cross-scale information. Specifically, we propose to utilize multi-scale dilated convolution~\cite{yu2015multi} to fuse the information together. For the low-level feature with the shape $\frac{H}{2^k}\times \frac{W}{2^k}\times 2^k C$, we apply dilated convolutions with stride $2^{3-k}$ and dilation from 1 to $2^{2-k}$. Then we concatenate all the acquired features together and fuse them with a linear layer to derive the cross-scale appearance feature of the $i$-th frame $\bm{C}_i$. In this way, the more fine-grained the feature, the more information it will provide.
        
        Afterward, $\bm{C}_0$ and $\bm{C}_1$ are fed into the hierarchical motion-appearance feature extractor composed of the Transformer block containing the inter-frame attention to extract motion features $\bm{M}_{i}$ and inter-frame appearance features $\bm{A}_i$.  Following the recent motion-aware methods~\cite{kong2022ifrnet,huang2020rife,park2021asymmetric,lu2022video}, we first utilize the acquired motion and appearance feature to estimate the bidirectional optical flows $\bm{F}$ and masks $\bm{O}$, then we use them to warp the inputs frame to $t$ and fuse together, as:
        % \begin{align}
        %     &\tilde{\bm{I}_t} = \bm{O}_0\odot\text{BW}\left(\bm{I}_0, \bm{F}_{t\rightarrow 0}\right) + \bm{O}_1\odot\text{BW}\left(\bm{I}_1, \bm{F}_{t\rightarrow 1}\right)\;,
        % \end{align}
        \begin{align}
        % \label{12}
            &\bm{\tilde{I}}_t = \bm{O}\odot\text{BW}\left(\bm{I}_0, \bm{F}_{t\rightarrow 0}\right) + \left(1-\bm{O}\right)\odot\text{BW}\left(\bm{I}_1, \bm{F}_{t\rightarrow 1}\right)\;,
        \end{align}
        where $\text{BW}$ is the backward warp operation~\cite{huang2020rife} and $\odot$ represents the Hadamard product. Finally, we further exploit the low-level features and inter-frame appearance features to refine the appearance of the fused frame $\tilde{\bm{I}_t}$ by the RefineNet:
        \begin{align}
            &\bm{\hat{I}}_t = \bm{\tilde{I}}_t + \text{RefineNet}\left(\bm{\tilde{I}}_t, \bm{L}, \bm{A}\right)\;.
        \end{align}

        Since the motion and appearance features already have enough information, only three convolution layers for estimating motion and  a simplified U-Net~\cite{ronneberger2015u} for the RefineNet is enough for excellent performance. The details of motion estimation and the RefineNet are provided in the \textbf{supplementary materials}. 
        
 %        \begin{figure}[t]
	% 	\begin{center}
	% 		%\fbox{\rule{0pt}{2in} \rule{0.9\linewidth}{0pt}}
	% 		\includegraphics[width=0.95\linewidth]{imgs/Method3.pdf}
	% 	\end{center}
	% 	\vspace{-0.1in}
	% 	\caption{Proposed cross-scale patch embedding. We utilize multi-scale dilated convolution~\cite{yu2015multi} to fuse the low-level information together to enhance the input information for inter-frame attention.}
	% 	\label{fig:cspe}
	% 	\vspace{-0.1in}
	% \end{figure}
 
        % \subsection{Cross-Scale Patch Embedding}
        % \label{sec:cspe}
        % Though the integrated CNN and Transformers design could relieve the overhead, it also caused a lack of fine-grained information in Transformers' input. To alleviate this problem, we reuse the low-level features derived by CNNs to complement the cross-scale information. As shown in \cref{fig:cspe}, we propose to utilize multi-scale dilated convolution~\cite{yu2015multi} to fuse the information together. For the low-level feature with the shape $\frac{H}{2^k}\times \frac{W}{2^k}\times 2^k C$, we apply dilated convolutions with stride $2^k$ and dilation from 1 to $2^{k-1}$. In this way, the receptive field keeps consistent between features with different scales, and the more fine-grained the feature, the more information it will provide. Finally, we concatenate all the features and perform a linear layer to fuse them together.

\section{Experiments}

\begin{table*}[t]
\renewcommand\arraystretch{1.15}
	\caption{Quantitative comparison among different benchmarks~(IE on Middlebury, PSNR/SSIM on other datasets). The best result and the second best are \first{boldfaced} and \second{underlined} respectively. ``Out of Memory'' is denoted as ``OOM'', and ``\Checkmark'' in ``Extra'' implies extra pre-trained models are used for training. ``\dag''~indicates the results obtained by ourselves, the rest of the results are copied from \cite{huang2020rife, reda2022film, lu2022video, hu2022many, kong2022ifrnet}. We use the V100 GPU for testing and follow the test procedure of \cite{huang2020rife} on Vimeo90K/UCF101/Middlebury, \cite{niklaus2020softmax} on Xiph, \cite{kong2022ifrnet} on SNU-FILM, respectively.  Note that we retested M2M on Xiph in order to be consistent with the procedure of \cite{niklaus2020softmax} for a fair comparison.}
        \vspace{-0.12in}
	\centering
	\resizebox{\textwidth}{!}{
		\begin{tabular}{lcccccccccc}
			\toprule 
			\multicolumn{1}{l}{\multirow{2}{*}{Method}} & 
			\multicolumn{1}{c}{\multirow{2}{*}{Extra}} &
			\multicolumn{1}{c}{\multirow{2}{*}{Vimeo90K}} & 
			\multicolumn{1}{c}{\multirow{2}{*}{UCF101}} & 
			\multicolumn{2}{c}{Xiph} & 
			\multicolumn{1}{c}{\multirow{2}{*}{M.B.}} & 
			\multicolumn{4}{c}{SNU-FILM}\\
			\cmidrule{5-6} \cmidrule{8-11}&  & & & 2K& 4K& & Easy & Medium & Hard & Extreme \\
			\midrule
			\multicolumn{2}{l}{\textbf{\textit{Two-Stage Training}}}  &  &  &  &  &  &  &  &  & \\ 
			BMBC~\cite{park2020bmbc} &-- & 35.01/0.9764 & 35.15/0.9689 & 32.82/0.928 & 31.19/0.880 & 2.04 & 39.90/0.9902 & 35.31/0.9774 & 29.33/0.9270 & 23.92/0.8432 \\ 
			ABME~\cite{park2021asymmetric} &-- & 36.18/0.9805 & 35.38/0.9698 & 36.53/\second{0.944} & 33.73/0.901 & 2.01 & 39.59/0.9901 & 35.77/0.9789 &  30.58/0.9364 &  25.42/0.8639 \\ 
			VFIFormer~\cite{lu2022video} & \Checkmark & \second{36.50}/\second{0.9816} &  \second{35.43}/\second{0.9700} & OOM\dag & OOM\dag &  1.82 &  \first{40.13}/\second{0.9907} & \first{36.09}/\second{0.9799} &  30.67/\second{0.9378} & 25.43/\second{0.8643}\\
			\midrule
			\multicolumn{2}{l}{\textbf{\textit{Single-Stage Training}}}  &  &  &  &  &  &  &  &  & \\ 
			ToFlow~\cite{baker2011database}&-- & 33.73/0.9682 & 34.58/0.9667 & 33.93/0.922 & 30.74/0.856 & 2.15 & 39.08/0.9890 & 34.39/0.9740 & 28.44/0.9180 & 23.39/0.8310 \\ 
			SepConv~\cite{niklaus2017video}&-- & 33.79/0.9702 & 34.78/0.9669 & 34.77/0.929 & 32.06/0.880 & 2.27 & 39.41/0.9900 & 34.97/0.9762 & 29.36/0.9253 & 24.31/0.8448 \\ 
			DAIN~\cite{bao2019depth}& \Checkmark & 34.71/0.9756 & 34.99/0.9683 &35.95/0.940 & 33.49/0.895 & 2.04 & 39.73/0.9902 & 35.46/0.9780 & 30.17/0.9335 & 25.09/0.8584 \\ 
			AdaCoF~\cite{lee2020adacof}&-- & 34.47/0.9730 & 34.90/0.9680 & 34.86/0.928 & 31.68/0.870 & 2.24 & 39.80/0.9900 & 35.05/0.9754 & 29.46/0.9244 & 24.31/0.8439 \\ 
			CAIN~\cite{choi2020channel}&-- & 34.65/0.9730 & 34.91/0.9690 & 35.21/0.937 & 32.56/0.901 & 2.28 & 39.89/0.9900 & 35.61/0.9776 & 29.90/0.9292 & 24.78/0.8507 \\ 
			SoftSplat~\cite{niklaus2020softmax}&  \Checkmark & 36.10/0.9802 & 35.39/0.9697 & \second{36.62}/\second{0.944}& 33.60/0.901 &  \first{1.81} & -- & -- & -- & -- \\ 
            M2M~\cite{hu2022many}& \Checkmark & 35.47/0.9778 & 35.28/0.9694 & 36.44/0.943\dag & 33.92/0.899\dag & 2.09\dag & 39.66/0.9904\dag & 35.74/0.9794\dag  & 30.30/0.9360\dag & 25.08/0.8604\dag \\ 
			IFRNet~\cite{kong2022ifrnet}& \Checkmark& 35.80/0.9794 & 35.29/0.9693 & 36.00/0.936\dag & 33.99/0.893\dag &  1.95 &  \second{40.03}/0.9905 & 35.94/0.9793 & 30.41/0.9358 & 25.05/0.8587 \\ 
            RIFE~\cite{huang2020rife}&-- & 35.61/0.9779 & 35.28/0.9690 & 36.19/0.938\dag & 33.76/0.894\dag &  1.96 &  39.80/0.9903\dag & 35.76/0.9787\dag & 30.36/0.9351\dag & 25.27/0.8601\dag \\ 
            \textbf{Ours-small}&-- & 36.07/0.9797 & 35.34/0.9696& 36.55/0.942 & \second{34.25}/\second{0.902} & 1.94 &  39.81/0.9906 & 35.88/0.9795 & \second{30.69}/0.9375 & \second{25.47}/0.8632  \\
            \textbf{Ours}&-- &  \first{36.64}/\first{0.9819} & \first{35.48}/\first{0.9701} & \first{36.90}/\first{0.945} & \first{34.67}/\first{0.907} & \first{1.81} & 39.98/\first{0.9910} & \first{36.09}/\first{0.9801} & \first{30.94}/\first{0.9392} & \first{25.69}/\first{0.8661}  \\
			\bottomrule 
	\end{tabular}}
	\label{table:quantitative1}
	\vspace{-0.10in}
\end{table*}
	
    \subsection{Datasets}
    Our model is evaluated on various datasets illustrated as follows: \textbf{1) Vimeo90K}~\cite{xue2019video}, which is composed of two subsets with a fixed resolution of $448\times 256$, i.e. the Triplet and Septuplet datasets.  \textbf{2) UCF101}~\cite{soomro2012ucf101}, which is related to human actions and contains 379 triplets with a resolution of $256\times 256$. \textbf{3) Middlebury}~\cite{baker2011database}, we use the OTHER set in Middlebury for testing, which contains images with a resolution around $640\times 480$. \textbf{4) SNU-FILM}~\cite{choi2020channel}, it contains 1,240 triplets with 1280x720 resolution, and is divided into four subsets with different levels of difficulty: Easy, Medium, Hard, and Extreme. \textbf{5) Xiph}~\cite{xphi1994}, following \cite{niklaus2020softmax}, we downsample and center-corp the original image to 2K resolution to get ``Xiph-2K'' and ``Xiph-4K''. \textbf{6) HD}~\cite{bao2019memc}, it contains 11 videos at three different resolutions of 544p, 720p and 1080p, and we follow the procedure of \cite{huang2020rife} to test arbitrary-timestep frame synthesis. \textbf{7) X4K1000FPS}~\cite{sim2021xvfi}, it is a 4K dataset proposed by \cite{sim2021xvfi}. We follow the test procedure of \cite{hu2022many}, performing arbitrary-timestep frame synthesis testing under both 4K and downsampled 2K resolutions.
    
    \begin{figure}[t]
		\begin{center}
			%\fbox{\rule{0pt}{2in} \rule{0.9\linewidth}{0pt}}
			\includegraphics[width=1\linewidth]{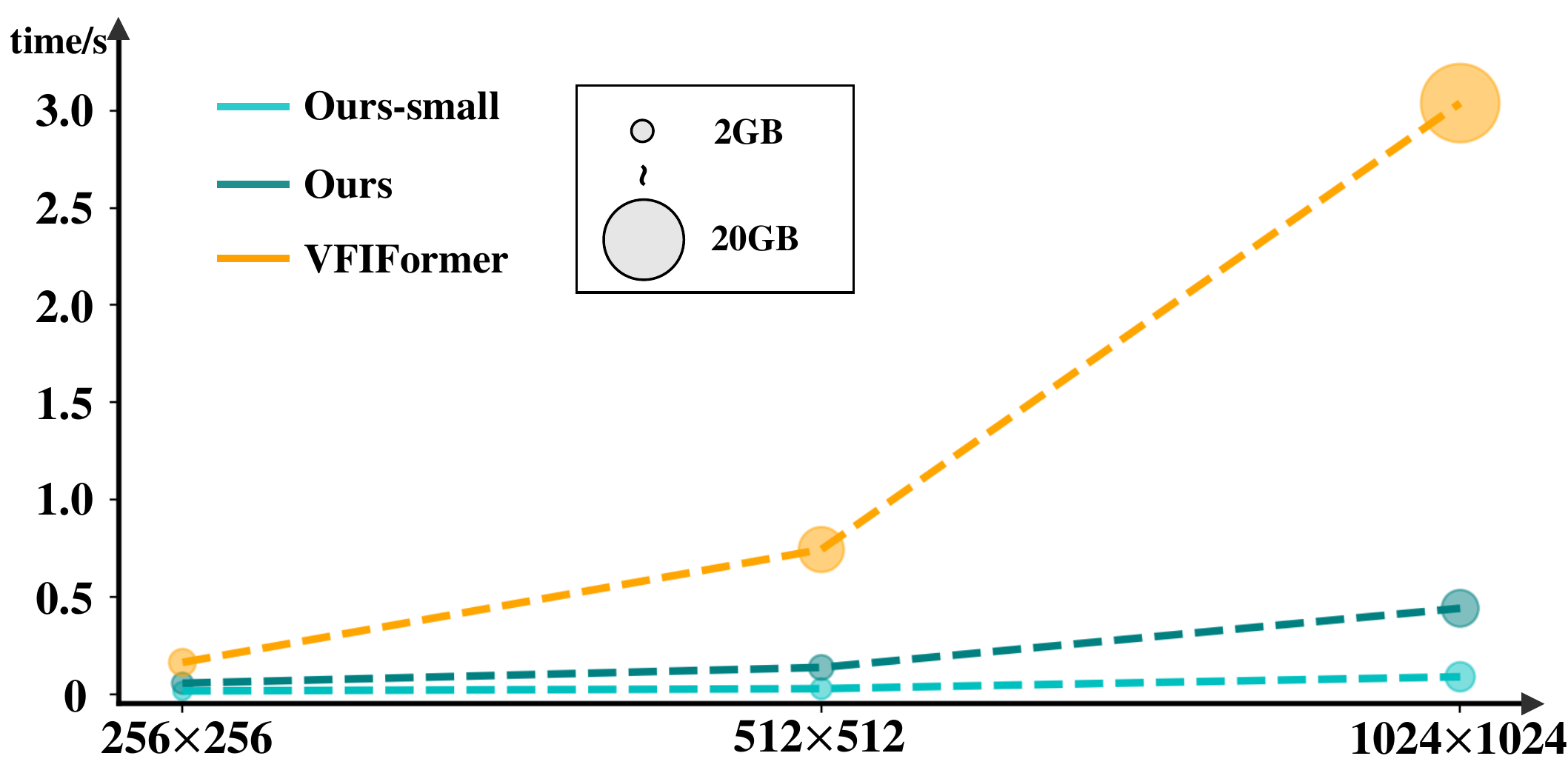}
		\end{center}
            \vspace{-0.2in}
		\caption{Comparison between our models and VFIFormer in terms of speed and memory usage at different input resolutions.}
		\label{fig:st}
		\vspace{-0.2in}
    \end{figure}
    
    \subsection{Implementation Details}
    \label{exp:td}
    \noindent
    \textbf{Model Configuration.} To show the scalable capability of our proposed module, we present two versions of our model: a time-friendly small model~(Ours-small) and a larger but more accurate model~(Ours). For the small model, the number of Transformer blocks at each stage ($N_1$ and $N_2$ in \cref{fig:overall}) is 2 and the initial channel number $C$ is 16. For the larger model, those are 4 and 32 respectively. We choose shifted window attention~\cite{liu2021swin} as the inter-frame attention and the window size is set to $7$. The remaining structures stay the same for both models. Following \cite{huang2020rife}, we apply the test-time argument to boost the performance of the larger. The original performance is provided in the ablation study.

    \noindent
    \textbf{Training Details.} 
    For fixed-timestep frame interpolation, we train our models on the triplet set of Vimeo90K~\cite{xue2019video}, in which $t=0.5$. We crop each frame to $256 \times 256$ patches and perform the random flip, time reversal, and rotation argumentation. The training batch size is set to 32. We choose AdamW~\cite{loshchilov2018fixing} as the optimizer with $\beta_1=0.9$, $\beta_2=0.999$ and weight decay $1e^{-4}$. We first warm up for 2000 steps to increase the learning rate to $2e^{-4}$ and then utilize cosine annealing~\cite{loshchilov2016sgdr} for 300 epochs to reduce the learning rate from $2e^{-4}$ to $2e^{-5}$. For arbitrary-timestep frame interpolation, we follow the same training procedure of~\cite{huang2020rife}, which randomly selects 3 frames from septuplet of Vimeo90K and calculated corresponding $t$. There is no change in the remaining settings. The training loss basically follows \cite{niklaus2020softmax,huang2020rife}, which is included in the supplementary file. 

    \begin{table}[t]
	\renewcommand\arraystretch{1.15}
	\setlength{\belowcaptionskip}{0pt}
	\caption{Quantitative comparison  for $4\times$ interpolation on HD and $8\times$ interpolation on XTest. We follow the test procedure of \cite{huang2020rife} on HD and \cite{hu2022many} on XTest. All notations are consistent with \cref{table:quantitative1}. All results except those marked with ``\dag'' are extracted from \cite{hu2022many, huang2020rife}.}
        \vspace{-0.12in}
	\resizebox{\columnwidth}{!}{
		\begin{tabular}{lccccc}
			\toprule 
			\multicolumn{1}{l}{Method} & 
			\multicolumn{1}{c}{HD(544p)} & 
			\multicolumn{1}{c}{HD(720p)} & 
			\multicolumn{1}{c}{HD(1080p)} & 
			\multicolumn{1}{c}{XTest-2K} &
			\multicolumn{1}{c}{XTest-4K}\\
			\midrule 
			DAIN~\cite{bao2019depth} & 22.17 & 30.25 & -- & 29.33 & 26.78  \\
			CAIN~\cite{choi2020channel} & 21.81 & 31.59 & 31.08 & 23.62 & 22.51\\
% 			BMBC~\cite{park2020bmbc} & 19.51 & 23.47 & OOM & & OOM\\ 
% 			DSepconv~\cite{cheng2020video} & 19.28 & 23.48 & OOM & & OOM \\
% 			CDFI~\cite{ding2021cdfi} & 21.85 & 29.28 & OOM & & OOM \\
% 			EDSC$_m$~\cite{cheng2021multiple} & 21.89 & 30.35 & 30.91 & & \\
			ABME~\cite{park2021asymmetric} & 22.46 &  31.43 & 33.22 & 30.65 & 30.16 \\ 
			RIFE$_m$~\cite{huang2020rife} & 22.95 & 31.87 & 34.25 & 31.43\dag & 30.58 \\
			IFRNet~\cite{kong2022ifrnet} & 22.01\dag & 31.85\dag & 33.19\dag & 31.53\dag & 30.46\dag \\
			M2M~\cite{hu2022many} & 22.31\dag & 31.94\dag & 33.45\dag & \second{32.13} & 30.88 \\
			\textbf{Ours-small} & \second{23.26} & \second{32.17} & \second{34.65} & 31.89 &  \second{30.89} \\
            \textbf{Ours} & \first{23.62} & \first{32.38} & \first{35.28} & \first{32.85} & \first{31.46} \\
			\bottomrule 
	\end{tabular}}
	\label{table:quantitative2}
    \vspace{-0.20in}
    \end{table}
    
    \subsection{Comparison with the State-of-the-Art Methods}
    To inspect the generalization ability of our proposed methods, we evaluate our model on diverse datasets and compared results with recent VFI approaches, which include: ToFlow~\cite{baker2011database}, SepConv~\cite{niklaus2017video}, AdaCoF~\cite{lee2020adacof}, CAIN~\cite{choi2020channel}, DAIN~\cite{bao2019depth}, BMBC~\cite{park2020bmbc}, ABME~\cite{park2021asymmetric}, IFRNet~\cite{kong2022ifrnet}, RIFE~\cite{huang2020rife}, SoftSplat~\cite{niklaus2020softmax}, 
    and VFIFormer~\cite{lu2022video}.

    \begin{figure*}[t]
		\begin{center}
			\includegraphics[width=0.85\linewidth]{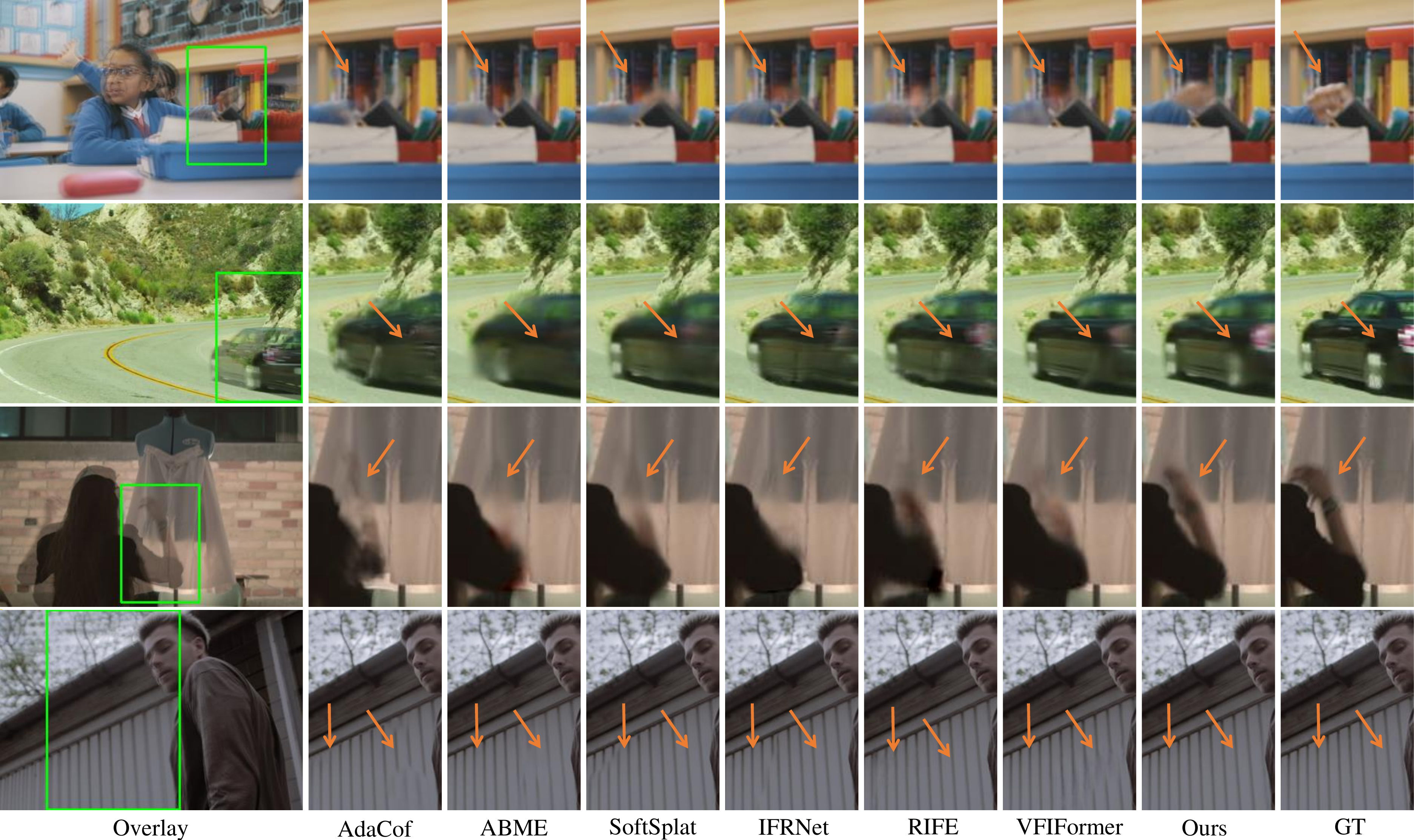}
		\end{center}
		\vspace{-0.20in}
		\caption{Visual comparison on Vimeo90K~\cite{xue2019video} triplet set. The position pointed by the arrow indicates where our model performs better.}
		\label{fig:exp1}
		\vspace{-0.15in}   
     \end{figure*}

    \noindent
    \textbf{Fixed Timestep Interpolation.} \cref{table:quantitative1} shows the results of fixed timestep interpolation ($t=0.5$) on various datasets. Our approach achieves state-of-the-art performance on almost all test sets except for the Easy set of SNU-FILM, which we attribute the reason to the fact that we did not apply inter-frame attention to the high-resolution features for a balance between performance and speed. As shown in \cref{fig:st}, as the input size increases, compared to the previous SOTA model, VFIFormer, our model dominates in terms of speed and memory usage, and still maintains better performance. Remarkably, our method has a more significant improvement on large motion datasets. Compared to the previous SOTA, our method has \textbf{0.28} dB and \textbf{0.68} dB improvements on the 2K and 4K sets of Xiph respectively as well as \textbf{0.27} dB and \textbf{0.26} dB improvements on Hard and Extreme sets of SNU-FILM respectively.
    
    \noindent
    \textbf{Arbitrary Timestep Interpolation.} Following \cite{huang2020rife}, we provide the results of multiple frame interpolation on HD benchmark~\cite{bao2019memc} and X4K1000FPS~\cite{sim2021xvfi}, as shown in \cref{table:quantitative2}. Thanks to the explicit motion features that can be used as cues for arbitrary-timestep interpolation, our approaches achieve the best performance on \textbf{all the test datasets}.
     \begin{table}[t]
    \setlength{\belowcaptionskip}{0pt}
    \renewcommand\arraystretch{1.15}
    \centering
    \caption{Ablation on the inter-frame attention. We use ``SFA'' to denote the single frame attention which only applies self-attention within a single frame, ``Mixed'' to denote the attention conducted within two frames together, and ``BCV'' to denote the bilateral cost volume proposed by \cite{park2020bmbc}.}
    \vspace{-0.12in}
    \resizebox{\columnwidth}{!}{
    \begin{tabular}{ll|cccc}
    \toprule
    Appearance & \multicolumn{1}{c|}{Motion} & Vimeo90K & Xiph-2K & Xiph-4K & Runtime \\
    \midrule
    SFA & \XSolidBrush & 35.54/0.977 & 36.26/0.939 & 33.36/0.895 & \textbf{26ms} \\
    IFA & \XSolidBrush & \textbf{36.02}/\textbf{0.980} & \textbf{36.49}/\textbf{0.942} & \textbf{34.20}/\textbf{0.902}& 27ms \\
    \midrule
    \multicolumn{2}{c|}{Mixed} & 35.54/0.978 & 35.98/0.939 & 33.88/0.899 & \textbf{26ms}\\
    SFA & BCV & 35.70/0.978 & 36.22/0.939 & 33.34/0.895 & 297ms \\
    IFA & IFA & \textbf{36.07}/\textbf{0.980} & \textbf{36.55}/\textbf{0.942} & \textbf{34.25}/\textbf{0.902} & 30ms \\
    \bottomrule
    \end{tabular}
    }
    \label{table:ablation1}
    \vspace{-0.25in}
    \end{table}
    
\begin{figure*}[t]
		\begin{center}
			\includegraphics[width=0.9\textwidth]{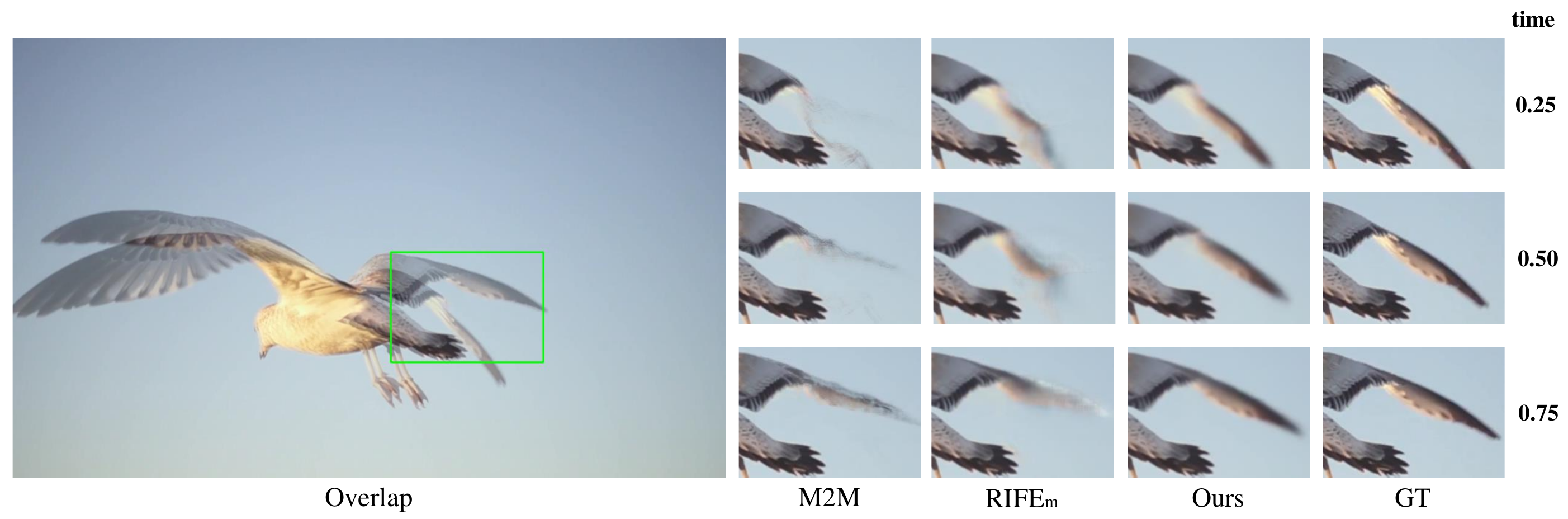}
		\end{center}
		\vspace{-0.20in}
		\caption{Visual comparison for multi-timestep generation selected from SNU-FILM~\cite{choi2020channel}.}
		\label{fig:exp2}
		\vspace{-0.20in}   
     \end{figure*}

    \noindent
    \textbf{Qualitative Comparison.} To underpin our quantitative results, we also give visual comparisons between our approaches and other VFI methods in intermediate and multi-frame generation respectively. As shown in \cref{fig:exp1}, compared to other methods, our model provides a superior estimation of the corresponding location of objects in the intermediate frames in the case of large motions and more favorable maintenance of texture information. Our model also exhibits better temporal consistency for complex motions in the multi-frame interpolation case, as shown in \cref{fig:exp2}.

    \subsection{Ablation Study}
    \label{exp:ab}
    In this section, we use the \textbf{small model~(Ours-small)} as the baseline to conduct ablation studies for investigating our proposed modules. The training settings are the same as \cref{exp:td} and we provide the test results of Vimeo90K and Xiph in order to observe the performance on both small- and large-motion datasets. We uniformly measure the time of processing a pair of 480p~($640 \times 480$) inputs for each model on the same device~(2080Ti), denoted as runtime.
    
     \noindent
     \textbf{Effect of the Inter-Frame Attention.} As the core operation of our proposed model, inter-frame attention~(IFA) can enhance the appearance information of each frame and extract bilateral motion information simultaneously. To verify its effectiveness, we replace IFA with different forms of attention as well as cost volume to extract appearance and motion information. As shown in \cref{table:ablation1}, when using only appearance information, the enhanced inter-frame appearance feature outperforms the single frame appearance feature substantially. When both appearance and motion information are used, our performance is further enhanced with only a slight increase in runtime.

    \begin{table}
    \centering
    \makebox[0pt][c]{\parbox{\columnwidth}{%
        \begin{minipage}[b]{\hsize}\centering
            \caption{Ablation on motion cues for arbitrary-timestep interpolation. ``$\bm{M}_t$'' indicates that motion features is used as cues and ``$+t$'' denotes directly input $t$ as cues.}
            \vspace{-0.12in}
            \resizebox{0.7\hsize}{!}{
                \begin{tabular}{lcccc}
                    \toprule
                    Cues & HD(720p) & XTest-2K & XTest-4K & Runtime\\
                    \midrule
                    \multicolumn{1}{c}{$+t$}  & 32.05 & 31.71 & 30.63 & \textbf{27ms} \\
                    \multicolumn{1}{c}{$\bm{M}_t$}  & \textbf{32.17} & \textbf{31.89} & \textbf{30.89} & 30ms\\
                    \bottomrule
                \end{tabular}
                }
            \label{table:ablation2}
        \end{minipage}
        % \hfill
        % \begin{minipage}[b]{0.48\hsize}\centering
        %     \caption{Efficiency analysis of multi-frame generation.}
        %     \vspace{-0.12in}
        %     \resizebox{\hsize}{!}{
        %         \begin{tabular}{lccc}
        %             \toprule
        %             Runtime & $8\times$ & $16\times$ & $32\times$ \\
        %             \midrule
        %             \multicolumn{1}{c}{original}  & 99 & 99 & 99 \\
        %             \multicolumn{1}{c}{shared}  & 99 & 99 & 99 \\
        %             \bottomrule
        %         \end{tabular}}
        %     \label{table:ablation3}
        % \end{minipage}%
    }}
    \vspace{-0.25in}
    \end{table}

     \noindent
     \textbf{Motion Cues for Arbitrary-Timestep Interpolation.} We use the motion feature extracted by inter-frame attention as the trigger to predict arbitrary-timestep frames. To verify its effectiveness, we compare it with the previous approaches which directly concatenate $t$ into the appearance feature as motion cues. As shown in \cref{table:ablation2}, using motion features as cues achieves better results on multiple datasets and maintains almost the same inference time.
     
        \begin{table}[t]
        \setlength{\belowcaptionskip}{0pt}
        \renewcommand\arraystretch{1.1}
        \centering
        \caption{Ablation on the scalable capability of Transformer blocks.}
        \vspace{-0.12in}
        \resizebox{\columnwidth}{!}{
        \begin{tabular}{cccccc}
        \toprule
        $N_1$\ /\ $N_2$ & C & Vimeo90K & Xiph-2K & Xiph-4K & Runtime \\
        \midrule
        2\ /\ 2 & 16 & 36.07/0.980 & 36.55/0.942 & 34.25/0.902 & \textbf{30ms}\\
        4\ /\ 4 & 16 & 36.21/0.980 & 36.61/0.943 & 34.31/0.902 & 39ms\\ 
        2\ /\ 2 & 32 & 36.43/0.981 & 36.70/0.943 & 34.51/0.905 & 66ms\\ 
        4\ /\ 4 & 32 & \textbf{36.50}/\textbf{0.981} & \textbf{36.74}/\textbf{0.944} & \textbf{34.55}/\textbf{0.906} & 78ms\\ 
        \bottomrule
        \end{tabular}
        }
        \label{table:ablation3}
        \vspace{-0.25in}
        \end{table}

     \noindent
     \textbf{Scalable Capability of Transformer Blocks.} As we mentioned before, the overall performance of the model can be controlled by simply adjusting the number and complexity of Transformer blocks. To confirm this, we double the number of Transformer blocks or their channels. As shown in \cref{table:ablation3}, both modifications improve the performance considerably. Since the increase in model complexity caused by the double of channel numbers is greater, the performance improvement is also relatively more noticeable.
    
     \noindent
     \textbf{Explore the Balance between Performance and Efficiency.} To alleviate the computational burden caused by Transformers, we adopt a hybrid CNNs/Transformers design. To explore the performance bounds, we replace the Transformer with CNNs or vice versa. As shown in \cref{table:ablation4}, using the Transformer only on the lowest scale features will significantly degrade the model's performance, and using it at higher scales will not improve the performance much while the computational overhead increases considerably. 

    % \noindent
    % \textbf{Limitations and Future Work.} Though a nontrivial improvement has been achieved by our proposed methods, there are still some limitations worth exploring. First, despite the integrated CNN and Transformer could relieve computational overhead, it also restricts motion extraction by inter-frame attention within high-resolution appearance features. A more sensible strategy for applying inter-frame attention to fine-grained features efficiently still needs to be discovered. Second, the input of our methods is restricted to two consecutive frames, which results in the inability to leverage information from multiple consecutive frames.  In future work, we will attempt to extend our approach to multi-frame inputs without introducing excessive overhead.
    \section{Limitations and Future Work}
    Though a nontrivial improvement has been achieved by our proposed methods, there are still some limitations worth exploring. First, despite the hybrid CNN and Transformer could relieve computational overhead, it also restricts motion extraction by inter-frame attention within high-resolution appearance features. Second, the input of our methods is restricted to two consecutive frames, which results in the inability to leverage information from multiple consecutive frames.  In future work, we will attempt to extend our approach to multi-frame inputs without introducing excessive overhead. Meanwhile, we will also investigate how to utilize the inter-frame attention in other fields that also need those two types of information, such as action recognition and action detection.
    \begin{table}[t]
    \setlength{\belowcaptionskip}{0pt}
    \renewcommand\arraystretch{1.1}
    \centering
    \caption{Ablation on diﬀerent hybrid CNNs/Transformers designs. ``C'' or ``T'' denotes we apply convolutional layers or Transformer blocks at the corresponding stage.}
    \vspace{-0.12in}
    \resizebox{\columnwidth}{!}{
    \begin{tabular}{cccccc}
    \toprule
    Architecture & Vimeo90K & Xiph-2K & Xiph-4K & Runtime \\
    \midrule
    C - C - C - C - T & 35.26/0.974 & 34.43/0.922 & 31.44/0.868 & \textbf{21ms}\\ 
    C - C - C - T - T & 36.07/0.980 & 36.55/0.942 & 34.25/0.902 & 30ms\\ 
    C - C - T - T - T & \textbf{36.10}/\textbf{0.980} & \textbf{36.58}/\textbf{0.943} & \textbf{34.29}/\textbf{0.903} & 44ms\\
    \bottomrule
    \end{tabular}
    }
    \label{table:ablation4}
    \vspace{-0.20in}
    \end{table}
   
    \section{Conclusion}
    In this work, we propose to exploit inter-frame attention for extracting motion and appearance information in video frame interpolation. In particular, we utilize the correlation information hidden within the attention map to simultaneously enhance the appearance information and model motion. Meanwhile, we devise an hybrid CNN and Transformer framework to achieve a better trade-off between performance and efficiency. Experiment results show that our proposed module achieves state-of-the-art performance on both fixed- and arbitrary-timestep interpolation and enjoys effectiveness compared with the previous SOTA method.

    \paragraph {\bf Acknowledgements.} {\small Thanks to the equal contributions of Yuhan Zhu and Haonan Wang. This work is supported by the National Key R$\&$D Program of China (No. 2022ZD0160900), the National Natural Science Foundation of China (No. 62076119, No. 61921006), the Fundamental Research Funds for the Central Universities (No. 020214380091), and the Collaborative Innovation Center of Novel Software Technology and Industrialization.}

%%%%%%%%% REFERENCES
{\small
\bibliographystyle{ieee_fullname}
\bibliography{egbib}
}
\clearpage

\section*{Appendix}

\section*{A. 1. Motion Estimation}
\begin{figure}[h]
		\centering
		\includegraphics[width=0.8\linewidth]{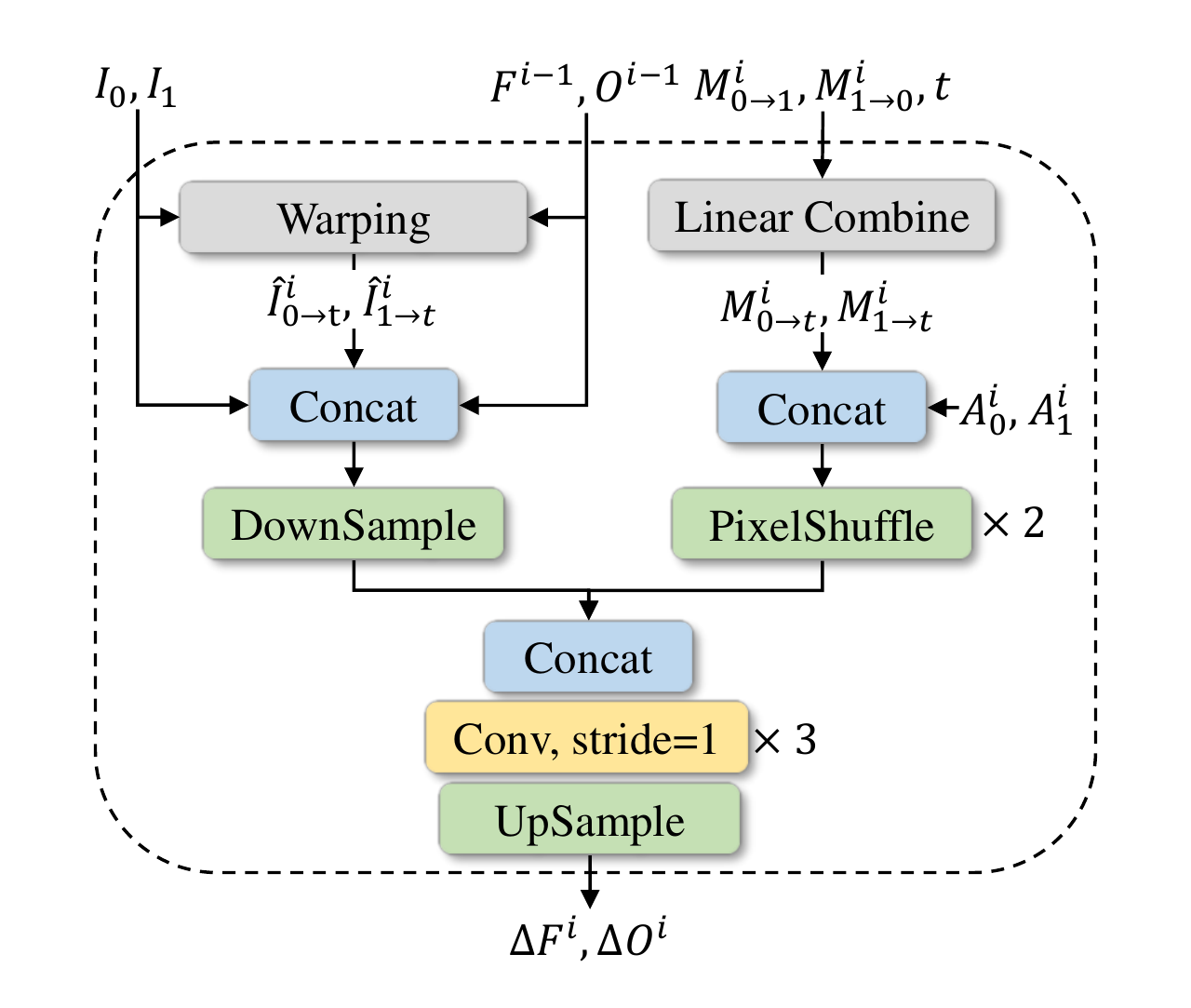}
		\caption{Pipeline of motion estimation.}
		\label{fig:flow}
		\vspace{-0.15in}
\end{figure}

	For motion estimation, we follow a similar design schema in RIFE~\cite{huang2020rife}, which directly utilizing simple convolutional layers to iteratively updates the optical flow $\bm{F}$ for backward warp and the fusion map $\bm{O}$. In contrast to RIFE, without as many as eleven convolutional layers in each iteration and extra privileged distillation, only three convolutional layers are needed for high-performance motion estimation due to the sufficient information contained in the extracted motion and appearance feature. 

	The pipeline of our motion estimation is shown in \cref{fig:flow}. For the motion and appearance features extracted at $i$-th Transformer stage, we first acquire $\bm{M}_{0\rightarrow t}^i$ and $\bm{M}_{1\rightarrow t}^i$ by linear scaling $\bm{M}_{0\rightarrow 1}^i$ and $\bm{M}_{0\rightarrow 1}^i$ with $t$. $\bm{M}_{0\rightarrow t}^i$ and $\bm{M}_{1\rightarrow t}^i$ are then concatenated with $\bm{A}_0^i$ and $\bm{A}_1^i$. Due to the resolution of motion and appearance features being quite low, we apply two PixelShuffle~\cite{shi2016real} with $r=2$ to quadruple the resolution of those features. To iteratively update $\bm{F}^{i-1}$ and $\bm{O}^{i-1}$ estimated in the previous stage, we also combine $\bm{F}^{i-1}$ and $\bm{O}^{i-1}$ with the warped original images as extra input to further boost the performance. Then the two stream inputs are concatenated together and fed to three convolution layers to generate the residual. The residual is upsampled by bilinear interpolation to the original resolution of inputs and added to the $\bm{F}^{i-1}$ and $\bm{O}^{i-1}$ to synthesize  motion at current stage:
	\begin{align}
            \bm{F}^{i} &= \bm{F}^{i-1} + \Delta \bm{F}^i,\\
            \bm{O}^{i} &= \bm{O}^{i-1} + \Delta \bm{O}^i
    \end{align}

\begin{figure}[t]
		\centering
		\includegraphics[width=0.7\linewidth]{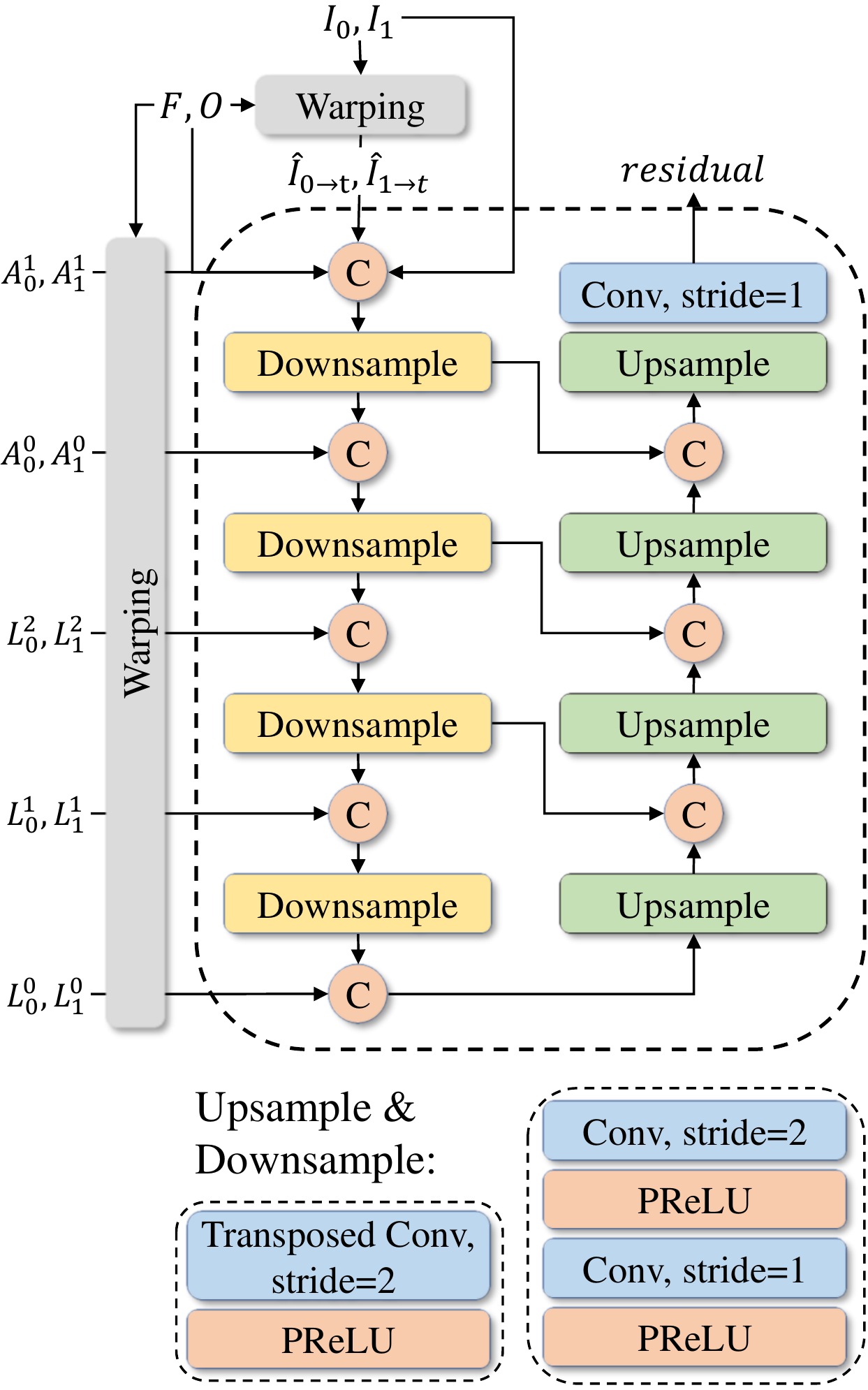}
		\caption{Structure of RefineNet.}
		\label{fig:refine}
		\vspace{-0.25in}
\end{figure}

\section*{A. 2. RefineNet}
We adopt a simplified U-Net~\cite{ronneberger2015u} architecture for refining the warped results $\tilde{\bm{I}}_t$ obtained with $\bm{F}$ and $\bm{O}$, as shown in \cref{fig:refine}. The only difference is that we add the acquired low-level features $L$ and inter-frame enhanced appearance features $A$ into the corresponding stage to provide additional appearance information for better appearance refinement.

\section*{A. 3. Loss Functions}The training loss is composed of two parts: warp loss and reconstruction loss. The warp loss is to directly supervise the result $\tilde{\bm{I}}_t$ obtained by warping and fusing inputs with $\bm{F}$ and $\bm{O}$, which implicitly supervise the motion estimation, as:
\begin{equation}
    \mathcal{L}_{warp}^{i}=f(\tilde{\bm{I}}_t^i,\bm{I}_t^{GT})\;,
\end{equation}
where $\mathcal{L}_{warp}^{i}$ represents the warps loss for $i$-th motion estimation, $\bm{I}_t^{GT}$ is the ground truth, $f$ is usually a pixel-wised loss. Following previous work \cite{niklaus2020softmax}, we employ the Laplacian loss, which denotes the $L_1$ loss between the Laplacian pyramids of the warped frame and the ground truth, as $f$. The reconstruction loss is to supervise the reconstruction quality of the final synthesized frame, as:
\begin{equation}
    \mathcal{L}_{rec}=f(\hat{\bm{I}}_t,\bm{I}_t^{GT})\;.
\end{equation}
The full loss function is defined as:
\begin{equation}
    \mathcal{L}=\mathcal{L}_{rec} + \lambda\sum_i \mathcal{L}_{warp}^{i}\;,
\end{equation}
where $\lambda$ is the loss weight for warp loss, we set $\lambda=0.5$ to maintain the balance between losses.

\section*{A. 4. Detailed Runtime/Memory Comparisons}
More comparisons, conducted on the 2080 Ti, are provided in \cref{table:compare}. Our method shows efficiency compared to high-performance models (VFIFormer and ABME), and Ours-small is comparable to real-time models (AdaCoF).
\begin{table}[h]
\setlength{\belowcaptionskip}{0pt}
\renewcommand\arraystretch{1.1}
\centering
\vspace{-0.12in}
\caption{More Runtime/Memory Comparisons.}
\vspace{-0.12in}
\resizebox{\columnwidth}{!}{
\begin{tabular}{c|ccc|cc}
\toprule
Input & Ours & VFIFormer & ABME & Ours-small & AdaCoF\\
\midrule
$256 \times 256$ & 56ms/1.49GB & 214ms/2.41GB & 84ms/1.50GB & 13ms/1.14GB & 6ms/1.19GB\\
$512 \times 512$ & 132ms/2.01GB & 892ms/6.13GB & 206ms/2.20GB & 25ms/1.42GB & 21ms/1.58GB\\ 
\bottomrule
\end{tabular}
}
\label{table:compare}
\label{t2}
\vspace{-0.25in}
\end{table}

\section*{A. 5. Affect of window size}
As shown in \cref{table:ablation5}, 7 is a decent choice for the attention window size.

\begin{table}[h]
        \setlength{\belowcaptionskip}{0pt}
        \renewcommand\arraystretch{1.1}
        \centering
        \vspace{-0.12in}
        \caption{Affect of window size.}
        \vspace{-0.12in}
        \resizebox{0.8\columnwidth}{!}{
        \begin{tabular}{c|ccc}
        \toprule
        Winow Size & Vimeo90k & Xiph-2k & Xiph-4k\\
        \midrule
        5 & 36.04/\textbf{0.9797} & 36.40/0.9418 & 34.21/0.9015\\
        7 & \textbf{36.07}/\textbf{0.9797} & \textbf{36.55}/\textbf{0.9421} & \textbf{34.25}/\textbf{0.9019}\\ 
        9 & 36.05/0.9795 & 36.53/0.9420 & 34.18/0.9011\\ 
        \bottomrule
        \end{tabular}
        }
        \label{table:ablation5}
        \label{t1}
        \vspace{-0.15in}
        \end{table}
\end{document}